\documentclass{article}

\usepackage[utf8]{inputenc}
\usepackage[T1]{fontenc}
\usepackage[english]{babel}
\usepackage[
  letterpaper,
  top=2cm,
  bottom=2cm,
  left=3cm,
  right=3cm,
  marginparwidth=1.75cm
]{geometry}

\usepackage[colorinlistoftodos]{todonotes}
\usepackage{comment}

\usetikzlibrary{arrows.meta}
\usepackage{graphicx}

\usepackage{amsmath,amssymb}
\usepackage{graphicx}
\usepackage{tabularx}
\usepackage{booktabs}
\usepackage{float}
\usepackage{xcolor}
\usepackage{tikz}
\usetikzlibrary{shapes.geometric,arrows.meta,positioning}

\usepackage[colorlinks=true,allcolors=blue]{hyperref}
\hypersetup{
  pdftitle={An Iterative LangGraph Agent for Text-to-SQL:
    Natural Language Access to the Chicago Crime Database},
  pdfauthor={Vigneshwar Ravi Rao, Rupesh Swarnakar,
    Fayeq Jeelani Syed}
}

\title{An Iterative LangGraph Agent for Text-to-SQL:\\
Natural Language Access to the Chicago Crime Database}

\author{%
  Vigneshwar Ravi Rao\textsuperscript{*},
  Rupesh Swarnakar\textsuperscript{*},
  and Fayeq Jeelani Syed\textsuperscript{\ensuremath{\dagger}}\\[6pt]
  {\small Luddy School of Informatics, Computing and Engineering, Indiana University, Indianapolis, Indiana, USA}\\[6pt]
  {\small \textsuperscript{*}These authors contributed equally.}\\
  {\small \textsuperscript{\ensuremath{\dagger}}Corresponding author:
  \texttt{faysyed@iu.edu}}
}

\date{}

\begin{document}
\maketitle

\begin{abstract}
\noindent\textbf{Purpose:} Adding components to a large language model (LLM) Text-to-SQL agent does not guarantee better answers. We evaluate this trade-off for a LangGraph agent over the Chicago Crime database.

\noindent\textbf{Methods:} Two agents sharing one model (\texttt{gpt-oss-120b}, temperature 0) were compared: a simple pipeline, and a full pipeline that adds a rule-based relevance checker and \texttt{EXPLAIN}-based validation with retry. Each was run with four cumulative prompt configurations, five times each, on 96 scored questions from a 100-question handcrafted benchmark (3,840 scored responses). Valid SQL rate, deterministic normalized result matching and LLM-judge-assisted execution accuracy are reported separately.

\noindent\textbf{Results:} Over all 96 questions with few-shot prompting, the simple agent reached 97.9\% valid SQL, 83.5\% normalized result matching and 91.5\% judge-assisted accuracy; the full agent reached 91.7\%, 78.3\% and 87.5\%. Validation and retry repaired all 160 observed first-pass validation failures, so every query the full agent produced executed; execution does not imply correctness. The relevance checker rejected the same eight of 96 in-scope questions in every run, imposing a 91.7\% ceiling on the full agent's overall valid SQL rate and accuracy. Synthesis-quality scores, confounded with run order, were not compared.

\noindent\textbf{Conclusion:} With explicit prompt rules and their one-shot and few-shot extensions, the simpler pipeline achieved higher judge-assisted query-result accuracy; with the bare prompt, the full pipeline performed better. At stages 2--4, an average of 1.5\% of the simple agent's queries failed to execute, leaving little for repair to recover. The agents differ in two components, so neither component's effect is isolated.
\end{abstract}


\medskip

\section{Introduction}

Relational databases are critical for operational intelligence, but their data remains largely inaccessible to non-technical stakeholders who lack SQL proficiency. This barrier forces business users to rely on engineering teams, creating operational bottlenecks and delaying actionable insights.

Large language models (LLMs) have made Text-to-SQL translation practical, but surveys of the field document persistent problems: questions mapped to the wrong schema elements, queries that do not execute, and ambiguous questions that admit more than one reasonable query \cite{qin2022survey, liu2025survey, shi2025survey}. Many systems also stop at query generation, leaving users to interpret raw result tables. A common engineering response is to wrap the query generator in additional components: input filters, query validators, retry loops and answer synthesizers. Each component targets a known failure mode, but each can also introduce failures of its own, so whether the combination improves the answers a user finally receives is an empirical question.

This paper reports an engineering evaluation of such an agent on a single realistic database. We built a Text-to-SQL agent with LangGraph \cite{langchain2024langgraph} for the Chicago Crime database and compared two versions of it that use the same model, schema and prompts: a simple pipeline, and a full pipeline that adds a relevance checker and a validate-and-retry loop. We distinguish four outcomes that are often reported together: (1) producing executable SQL, (2) returning the correct result, (3) summarizing that result accurately, and (4) preserving end-to-end performance once relevance filtering and retry are added.

\subsection{Related Work}

Text-to-SQL research has advanced through schema-aware encoding and in-context learning, benchmarked on datasets such as Spider \cite{yu2018spider} and models such as RAT-SQL \cite{ratsql2020}.

DIN-SQL \cite{pourreza2023din} decomposes Text-to-SQL generation into subtasks and incorporates self-correction, while DAIL-SQL \cite{gao2024dail} focuses on selecting and organizing in-context examples. Chain-of-thought prompting \cite{wei2022cot} provides a general approach to eliciting intermediate reasoning steps. Both studies evaluate GPT-4 \cite{openai2023gpt4}, and the DAIL-SQL study also investigates open-source models. SQL-specialized models such as SQLCoder \cite{defog2023sqlcoder} provide open-weight alternatives. MAC-SQL \cite{wang2023macsql} uses collaborating agents to generate SQL and refine erroneous queries using execution feedback. This use of feedback is conceptually related to ReAct's interleaving of reasoning and actions \cite{yao2023react}. BIRD \cite{li2023bird} extends evaluation to larger databases with challenges such as noisy contents, external knowledge requirements and query efficiency. Katsogiannis-Meimarakis and Koutrika \cite{katsogiannis2023survey} provide a broader survey of deep-learning approaches to Text-to-SQL.

Our agent draws on two of these ideas, execution-guided self-correction and fixed in-context examples, and adds a step that summarizes results in plain English. It is not a new architecture, and we make no claim of state-of-the-art performance. The contributions are empirical:
\begin{enumerate}
  \item a controlled comparison of two LangGraph agents that differ in relevance filtering and in validation with retry, across four prompt configurations with five repeated runs each;
  \item an evaluation that separates executable SQL, correct results and answer quality, and reports deterministic and LLM-judge-assisted scoring separately;
  \item a failure analysis showing that the retry loop repaired every observed first-pass validation failure while a deterministic relevance rule rejected eight in-scope questions, together with a description of a number-copying error in answer synthesis whose frequency changed with run order; and
  \item a reproducible database snapshot, benchmark and code.
\end{enumerate}

\section{Materials and Methods}

\subsection{Dataset and Data Processing}

The \textbf{Chicago Crime Dataset} (\href{https://data.cityofchicago.org/Public-Safety/Crimes-2001-to-Present/ijzp-q8t2}{Chicago Data Portal}) publishes incident records from 2001 to the present with 22 source attributes covering categorical (crime type, location description, beat/ward/district), temporal (date, year), geospatial (latitude/longitude, coordinates), and metadata (case number, arrest, domestic) fields. Data is ingested weekly via an Airflow ETL pipeline into a Dockerized PostgreSQL instance. Preprocessing standardized column names to \texttt{UPPERCASE\allowbreak\_SNAKE} format (for example, the source date field is stored as \texttt{INCIDENT\_DATE}), cast string timestamps to PostgreSQL \texttt{TIMESTAMP}, and cast coordinate fields to \texttt{FLOAT8}. The resulting fact table used in all experiments has 38 columns: the renamed source attributes and additional derived columns \texttt{CRIME\_PRIMARY\_TYPE\_CLEAN}, \texttt{CRIME\_DESCRIPTION\_CLEAN}, \texttt{CRIME\_TYPE\_KEY}, \texttt{CRIME\_OFFENSE\_LABEL}, \texttt{CRIME\_CATEGORY}, \texttt{CRIME\_OFFENSE\_GROUP}, \texttt{CRIME\_OFFENSE\_SUBTYPE}, \texttt{CRIME\_SEVERITY\_LEVEL}, \texttt{CRIME\_ \\
FBI\_CLASSIFICATION}, \texttt{CRIME\_CLEARANCE\_BUCKET}, \texttt{VIOLENT\_CRIME\_FLAG}, \texttt{PROPERTY\_CRIME\_FLAG}, \texttt{DRUG\_CRIME\_FLAG}, \texttt{WEAPON\_CRIME\_FLAG}, \texttt{LOCATION\_RISK\_CATEGORY}, \texttt{PROCESS\_DATE}. Because PostgreSQL folds unquoted identifiers to lower case, these upper-case names must be double-quoted in every query; the stage-2 prompt states this rule explicitly, and omitted quotes were the main cause of failed queries with the bare stage-1 prompt (Section~\ref{sec:results}). Null values were handled per column group as shown in Table~\ref{tab:null_handling}. Weekly Airflow schema validation confirms structural consistency before agent interactions.

\begin{table}[H]
\centering
\small
\begin{tabularx}{\textwidth}{lXXX}
\toprule[1pt]
\textbf{Column Group} & \textbf{Specific Columns} & \textbf{Handling Strategy} & \textbf{Rationale} \\
\midrule
\textbf{Categorical} & \texttt{CRIME\_TYPE}, \texttt{DESCRIPTION}, \texttt{LOCATION\_DESCRIPTION} & Replaced with \texttt{OTHER OFFENSE} & Permits downstream \texttt{COALESCE} operations in \texttt{SQL} \\
\midrule
\textbf{Geospatial} & \texttt{LATITUDE}, \texttt{LONGITUDE}, \texttt{X\_COORDINATE}, \texttt{Y\_COORDINATE} & Taken from a matching \texttt{BLOCK} where available; otherwise replaced with the mean for the \texttt{BEAT} & Allows spatial queries to filter records \\
\midrule
\textbf{Temporal} & \texttt{DATE}, \texttt{UPDATED\_ON}, \texttt{YEAR} & Missing \texttt{DATE} replaced with the most frequent date in the same \texttt{YEAR} & Malformed dates trigger query failures; ensures temporal consistency \\
\midrule
\textbf{Logical} & \texttt{ARREST}, \texttt{DOMESTIC} & Marked as ``Unknown''. Currently there are no Unknown values, since the source comes with all values filled & Distinguishes unknown status from a recorded value \\
\bottomrule[1pt]
\end{tabularx}
\caption{Null value handling strategy by column group. Column names are those of the source data before renaming. Imputed categories, coordinates and dates remain in the snapshot, so counts by category, location or date include imputed values.}
\label{tab:null_handling}
\end{table}

\subsection{Problem Formulation}

The task is \textbf{Text-to-SQL translation}: given a natural language question from a non-technical user, produce a syntactically valid PostgreSQL \texttt{SELECT} statement, execute it, and return a plain-English summary structured as \texttt{\{sql\_query: str, results: List[Tuple], answer: str\}}. Three protections against data modification are implemented: both agents connect through a read-only database role; the stage-2 to stage-4 prompts instruct the model never to write \texttt{INSERT}, \texttt{UPDATE}, \texttt{DELETE}, \texttt{DROP} or \texttt{ALTER} statements; and \texttt{agent\_2}'s validator additionally rejects queries containing these keywords. Only the database role applies to \texttt{agent\_1} at every stage. Scope is restricted to the Chicago Crime database. Supported query patterns include \texttt{JOIN}, \texttt{GROUP BY}, \texttt{HAVING}, \texttt{ORDER BY}, \texttt{LIMIT}, and subqueries. The schema is assumed stable between weekly validations, and a 7-day data refresh lag applies to real-time queries.

\subsection{Model / System Architecture}

\subsubsection{System Workflow}

Daily data ingestion via Airflow populates a Dockerized PostgreSQL database. The LangGraph AI Agent receives user questions, queries the database, and returns natural language answers. Figure~\ref{fig:overall_workflow} shows the overall system architecture; Figure~\ref{fig:two_agents} details the internal graph of the two agents compared in this work.

\begin{figure}[H]
\centering
\begin{tikzpicture}[
    auto,
    process/.style={rectangle, rounded corners, draw=black, thick, fill=blue!5,
                    text width=13em, minimum height=3em, text centered, inner sep=5pt},
    io/.style={rectangle, draw=black, text width=8em, minimum height=2.5em, text centered, inner sep=5pt},
    arrow/.style={thick, ->, >=stealth}
]
\node (ingestion) [process] {Daily Data Ingestion \\ (Airflow)};
\node (postgres)  [process, below=1.2cm of ingestion] {PostgreSQL Database \\ (Docker Container)};
\node (ai_agent)  [process, below=1.2cm of postgres]  {LangGraph AI Agent};
\node (question)  [io, left=1cm of ai_agent]  {User Question};
\node (answer)    [io, right=1cm of ai_agent] {Natural Language Answer};
\draw [arrow] (ingestion) -- (postgres);
\draw [arrow] (postgres)  -- (ai_agent);
\draw [arrow] (question)  -- (ai_agent);
\draw [arrow] (ai_agent)  -- (answer);
\end{tikzpicture}
\caption{Overall system architecture. Apache Airflow ingests the Chicago Crime data weekly into a PostgreSQL database running in Docker; the LangGraph agent receives a user question, queries the database and returns a plain-English answer.}
\label{fig:overall_workflow}
\end{figure}

\paragraph{Reproducible database container.} To let others reproduce the benchmark results, we built a Docker container that runs PostgreSQL~15 with the complete database, all 8,643,493 incident records, the same data used to run every experiment in this paper. The published image (\texttt{vigneshwarr/chicago-crime-db}) has the data already loaded, so it starts in seconds and runs unchanged on Windows, macOS and Linux with Docker. Because the production database is refreshed weekly, the image is a fixed snapshot: it holds every incident from 2001-01-01 through 2026-09-29, exported on 2026-10-10. The agents connect through a read-only database role, so even \texttt{agent\_1}, which has no query validation, cannot modify the data. The benchmark code, the 100 questions with reference SQL, and the build files for the image are in the project repository (\texttt{github.com/Vigneshwarr3/chicago-text2sql}).

\begin{table}[H]
\centering
\small
\begin{tabular}{llrl}
\toprule
Table & Role & Rows & Key columns \\
\midrule
\texttt{crime\_incidents}      & fact table  & 8,643,493 & 38 columns, e.g.\ \texttt{INCIDENT\_DATE}, \texttt{CRIME\_TYPE}, \texttt{ARREST} \\
\texttt{district\_names}       & lookup      & 23        & \texttt{district\_id}, \texttt{district\_name} \\
\texttt{community\_area\_names} & lookup      & 77        & \texttt{community\_area\_id}, \texttt{community\_area\_name} \\
\texttt{ward\_names}           & lookup      & 50        & \texttt{ward\_id}, \texttt{ward\_name} \\
\bottomrule
\end{tabular}
\caption{Tables in the benchmark database snapshot (Docker image \texttt{vigneshwarr/chicago-crime-db}; incidents from 2001-01-01 to 2026-09-29). The fact table has 22 source attributes and 16 derived columns; the lookup tables map district, community-area and ward codes to names.}
\label{tab:db_snapshot}
\end{table}

\subsubsection{The Two Agents}

To measure what the safeguards in the agent contribute, we implement it twice as a LangGraph \texttt{StateGraph}: a simple pipeline, \texttt{agent\_1}, and the full pipeline, \texttt{agent\_2} (Figure~\ref{fig:two_agents}). Both use the same language model (\texttt{gpt-oss-120b} served by Groq, temperature 0), the same database schema (the \texttt{crime\_incidents} fact table and the \texttt{district\_names}, \texttt{community\_area\_names} and \texttt{ward\_names} lookup tables), the same query-generation node, whose system prompt is the only component varied across our four prompt stages, and the same answer synthesizer, which summarizes the first ten result rows. They differ only in the safeguards placed around the query generator (Table~\ref{tab:agent_components}).

\paragraph{\texttt{agent\_1}: simple pipeline.} The graph is a straight line: the schema fetcher reads the table definitions, the query generator asks the model for a SQL query, the query runner executes it as generated, and the answer synthesizer turns the returned rows into a plain-English answer. There is no relevance check, no validation and no retry. A query that fails to execute therefore ends the run with an error and no usable result.

\paragraph{\texttt{agent\_2}: full pipeline.} \texttt{agent\_2} adds two safeguards to the same chain. (i)~A \emph{relevance checker} at the entry point is a rule-based filter: a question must contain a data-related word and an action verb (for example ``how many'', ``list'', ``top''), and questions that match out-of-scope patterns (arithmetic, definitions, ``who/when/why'' questions) are rejected. No model call is used to decide relevance. For a rejected question no query is generated, and the run ends with a short out-of-scope reply,  the decision to reject is made by the rules alone. (ii)~A \emph{query validator} sits between the generator and the runner. It rejects statements containing data-modifying keywords (\texttt{UPDATE}, \texttt{DELETE}, \texttt{DROP}, \texttt{INSERT}, \texttt{ALTER}) and checks the remaining query with PostgreSQL's \texttt{EXPLAIN}. When validation fails, the error message is passed back to the query generator, which tries again. The generator is called at most three times per question in total (the initial attempt and at most two retries); if all three attempts fail, the agent returns a fixed message asking the user to rephrase the question.


\definecolor{lgnode}{HTML}{FAF9FF}
\definecolor{lgborder}{HTML}{000000}
\definecolor{lgend}{HTML}{F0ECFA}

\tikzset{
  lgbox/.style={
    rectangle,
    rounded corners=2pt,
    draw=lgborder,
    fill=lgnode,
    line width=0.6pt,
    minimum height=9mm,
    minimum width=36mm,
    align=center,
    font=\small
  },
  lgterm/.style={
    rectangle,
    rounded corners=12pt,
    draw=lgborder,
    fill=lgnode,
    line width=0.6pt,
    minimum height=8mm,
    minimum width=23mm,
    align=center,
    font=\small
  },
  lgendnode/.style={
    lgterm,
    fill=lgend
  },
  lgedge/.style={
    -{Stealth[length=2.2mm]},
    line width=0.6pt,
    draw=black!75,
    shorten >=0.5pt
  },
  lgcond/.style={
    lgedge,
    densely dashed
  },
  lglabel/.style={
    font=\scriptsize,
    text=black,
    fill=white,
    inner sep=1.5pt,
    align=center
  }
}

\begin{figure}[H]
\centering
\begin{tikzpicture}

  \node[lgterm]    (s1)  at (0,-1.45)  {\_\_start\_\_};
  \node[lgbox]     (sf1) at (0,-3.20)  {schema\_fetcher};
  \node[lgbox]     (qg1) at (0,-4.95)  {query\_generator};
  \node[lgbox]     (qr1) at (0,-6.70)  {query\_runner};
  \node[lgbox]     (as1) at (0,-8.45)  {answer\_synthesizer};
  \node[lgendnode] (e1)  at (0,-10.20) {\_\_end\_\_};

  \draw[lgedge] (s1.south)  -- (sf1.north);
  \draw[lgedge] (sf1.south) -- (qg1.north);
  \draw[lgedge] (qg1.south) -- (qr1.north);
  \draw[lgedge] (qr1.south) -- (as1.north);
  \draw[lgedge] (as1.south) -- (e1.north);

  \node[font=\small] at (0,-12.45) {(a) \texttt{agent\_1}};

  \node[lgterm]    (s2)  at (7,0)      {\_\_start\_\_};
  \node[lgbox]     (rc)  at (7,-1.60)  {relevance\_checker};
  \node[lgbox]     (sf2) at (7,-3.20)  {schema\_fetcher};
  \node[lgbox]     (qg2) at (7,-4.80)  {query\_generator};
  \node[lgbox]     (qv)  at (7,-6.80)  {query\_validator};
  \node[lgbox]     (qr2) at (7,-8.40)  {query\_runner};
  \node[lgbox]     (as2) at (7,-10.00) {answer\_synthesizer};
  \node[lgendnode] (e2)  at (7,-11.60) {\_\_end\_\_};

  \draw[lgedge] (s2.south) -- (rc.north);

  \draw[lgcond]
    (rc.south)
    -- node[lglabel, right=1pt] {relevant}
    (sf2.north);

  \draw[lgcond]
    (rc.west)
    -- (4.45,-1.60)
    -- (4.45,-11.60)
    -- (e2.west);

  \node[lglabel] at (4.45,-6.60) {out of\\scope};

  \draw[lgedge] (sf2.south) -- (qg2.north);

  \draw[lgedge]
    ([xshift=-7mm]qg2.south)
    -- node[lglabel, left=1pt] {validate}
    ([xshift=-7mm]qv.north);

  \draw[lgcond]
    ([xshift=7mm]qv.north)
    -- node[lglabel, right=1pt] {retry}
    ([xshift=7mm]qg2.south);

  \draw[lgcond]
    (qv.south)
    -- node[lglabel, right=1pt] {valid}
    (qr2.north);

  \draw[lgcond]
    (qv.east)
    -- (9.55,-6.80)
    -- (9.55,-10.00)
    -- (as2.east);

  \node[lglabel] at (9.55,-8.40) {3 attempts\\failed};

  \draw[lgedge] (qr2.south) -- (as2.north);
  \draw[lgedge] (as2.south) -- (e2.north);

  \node[font=\small] at (7,-12.45) {(b) \texttt{agent\_2}};

\end{tikzpicture}

\caption{Internal \texttt{StateGraph} architectures of the two
LangGraph agents: (a)~\texttt{agent\_1}, the simple pipeline;
and (b)~\texttt{agent\_2}, with relevance checking, validation
and retry. Solid arrows indicate unconditional transitions,
and dashed arrows indicate conditional transitions.
Out-of-scope questions terminate at the relevance checker.
The query validator routes valid queries to execution,
returns failed queries to the generator for retry, or routes
the workflow to the answer synthesizer to request rephrasing
after three failed attempts.}
\label{fig:two_agents}
\end{figure}

\begin{table}[H]
\centering
\begin{tabular}{lcc}
\toprule
Component & \texttt{agent\_1} & \texttt{agent\_2} \\
\midrule
Schema fetcher             & yes & yes \\
Query generator            & yes & yes \\
Query runner               & yes & yes \\
Answer synthesizer         & yes & yes \\
Relevance checker          & no  & yes \\
Query validator & no  & yes \\
Retry (at most 3 attempts in total)   & no  & yes \\
\bottomrule
\end{tabular}
\caption{Components of the two agents. All other settings (model, schema, prompts, answer synthesizer) are identical. Because the agents differ in two components, differences in their results are a joint effect of both.}
\label{tab:agent_components}
\end{table}

Because the two graphs differ in two components at once, the relevance checker and the validate-and-retry loop, any difference in their results is a joint effect of both, not of either component alone.

\subsubsection{Node Implementation}

Both agents are implemented as a LangGraph \texttt{StateGraph} whose shared state carries \texttt{question}, \texttt{sql\_query}, \texttt{schema}, \texttt{results}, \texttt{answer}, \texttt{error\_log}, \texttt{retry\_count}, and \texttt{is\_relevant}. \texttt{agent\_2} uses all six nodes in Table~\ref{tab:nodes}; \texttt{agent\_1} uses nodes 2, 3, 5 and 6.

\begin{table}[h]
\centering
\small
\begin{tabularx}{\textwidth}{lXX}
\toprule[1pt]
\textbf{Node} & \textbf{Purpose} & \textbf{Key Implementation Detail} \\
\midrule
1. Relevance Checker & Filters non-database questions (\texttt{agent\_2} only) & Regular-expression and keyword rule; outputs \texttt{is\_relevant} flag; routes to END or Schema Fetcher \\
\midrule
2. Schema Fetcher & Retrieves live DB schema & \texttt{SQLDatabase.get\_table\_info()} \\
\midrule
3. Query Generator & LLM generates SQL & System prompt varied by prompt stage (Table~\ref{tab:prompt_stages}); fixed (non-retrieved) examples at stages 3 and 4; temp=0, max\_tokens=500 \\
\midrule
4. Query Validator & Dry-runs SQL (\texttt{agent\_2} only) & \texttt{EXPLAIN} + data-modification keyword check; increments retry counter (at most 3 generation attempts in total) \\
\midrule
5. Query Runner & Executes SQL & \texttt{db.run(fetch="all")}; the full result is returned \\
\midrule
6. Answer Synthesizer & Narrates results in plain English & LLM summarizes the first 10 result rows; returns a fixed message when no query could be produced \\
\bottomrule[1pt]
\end{tabularx}
\caption{StateGraph nodes: functions and implementation details. \texttt{agent\_2} uses all six nodes; \texttt{agent\_1} uses nodes 2, 3, 5 and 6.}
\label{tab:nodes}
\end{table}

\noindent\textbf{Conditional edges (\texttt{agent\_2}):} The Relevance Checker routes to the Schema Fetcher (relevant) or END (irrelevant). The Query Validator returns to the Query Generator on failure; after the third failed attempt it routes to the Answer Synthesizer, which returns the fixed rephrasing message.

\paragraph{Three distinct row limits.} Three settings concern the number of result rows and should not be confused. (i)~The SQL itself: the stage-2 to stage-4 prompts instruct the model not to add a \texttt{LIMIT} clause unless the question asks for a top-$N$ or a specific number of rows (rule~8 in Figure~\ref{fig:prompt_stage2}); the stage-1 prompt gives no instruction. (ii)~Answer synthesis: the synthesizer receives only the first ten rows of the result. (iii)~Evaluation: results are compared on at most their first 1,000 rows (Section~\ref{sec:evaluation_framework}).

\subsubsection{Tools and Frameworks}

\begin{table}[H]
\centering
\begin{tabularx}{\textwidth}{lXX}
\toprule[1pt]
\textbf{Component} & \textbf{Technology} & \textbf{Rationale} \\
\midrule
\textbf{LLM} & Groq API (openai/gpt-oss-120b) & Hosted inference for an open-weight model \\
\midrule
\textbf{Agentic Framework} & LangGraph & Explicit state management; transparent execution flow; conditional routing \\
\midrule
\textbf{Database Driver} & LangChain SQLDatabase + SQLAlchemy & Abstraction over PostgreSQL dialect; built-in schema introspection \\
\midrule
\textbf{Database} & PostgreSQL 15 for all experiments (the production pipeline is described as compatible with version 14 or later) & Supports the query patterns in the benchmark \\
\midrule
\textbf{Containerization} & Docker & Environment reproducibility; isolation \\
\midrule
\textbf{Data Pipeline} & Apache Airflow & Scheduled weekly ingestion; automatic error handling; audit trail \\
\midrule
\textbf{Language} & Python (code requires 3.10 or newer, we ran on Python 3.11.14) & Mature ML ecosystem; LangChain/LangGraph native support \\
\bottomrule[1pt]
\end{tabularx}
\caption{Tools and frameworks used in the system.}
\label{tab:tools_frameworks}
\end{table}

\subsection{Prompt Configurations}

No model fine-tuning was performed; the system prompt of the query-generation node was the only element varied. We organized it into four cumulative stages, each adding one kind of guidance to the previous one (Table~\ref{tab:prompt_stages}); every other component of the pipeline, including the model, the schema shown to it and the answer synthesizer, was identical across stages. The relevance checker and the \texttt{EXPLAIN}-based retry loop of \texttt{agent\_2} are changes to the graph, not to the prompt, and are described in the previous subsection.

\begin{table}[H]
\centering
\small
\begin{tabularx}{\textwidth}{lXX}
\toprule[1pt]
\textbf{Stage} & \textbf{Prompt content} & \textbf{What it adds} \\
\midrule
1. Simple & One sentence: ``You are a SQL assistant. Write a SQL query for the user's question.'' & Baseline; no database knowledge beyond the schema in the user message \\
\midrule
2. Rules & Stage~1 replaced by a role statement, the foreign-key relationships to the lookup tables, and ten numbered rules (Figure~\ref{fig:prompt_stage2}) & PostgreSQL conventions: quoting of upper-case identifiers and string values, boolean and date handling, JOINs on names, CTEs and window functions, percentage formula, no \texttt{LIMIT} unless asked, read-only queries, output format \\
\midrule
3. One-shot & Stage~2 followed by one worked question--SQL pair (Figure~\ref{fig:prompt_stage3}) & A concrete example of the expected output style \\
\midrule
4. Few-shot & Stage~2 followed by four worked question--SQL pairs (Figure~\ref{fig:prompt_stage4}) & Examples covering counting, top-$N$ ranking, boolean filters and grouped aggregation with \texttt{HAVING} \\
\bottomrule[1pt]
\end{tabularx}
\caption{The four prompt stages. Stages 3 and 4 append examples to the stage~2 prompt unchanged.}
\label{tab:prompt_stages}
\end{table}

\paragraph{Stage 1: simple.} The bare prompt gives the model nothing beyond the table definitions supplied with each question. Its main weakness is identifier quoting: the schema uses upper-case column names, which PostgreSQL folds to lower case unless they are double-quoted. In our runs, 33\% of the queries \texttt{agent\_1} generated at this stage failed to execute, mostly with errors such as \texttt{column "incident\_date" does not exist}.

\paragraph{Stage 2: explicit rules.} The prompt states the database relationships (district, community area and ward codes in \texttt{crime\_incidents} map to names in the lookup tables) and ten rules that address the recurring failure modes: quoting identifiers and string values, comparing boolean columns directly, using half-open date ranges, joining lookup tables instead of guessing numeric codes, using CTEs and explicit window clauses for multi-step logic, computing percentages safely, and refusing non-\texttt{SELECT} requests. This stage produced the largest improvement in query validity.

\paragraph{Stages 3 and 4: examples.} Stage 3 appends one worked example and stage 4 appends four. The examples are the reference SQL of four questions of the evaluation set that are reserved for this purpose and never scored, so the model is never shown the answer to a question it is evaluated on.

\begin{figure}[H]
\centering
\fbox{\parbox{0.94\linewidth}{\ttfamily\scriptsize\raggedright
    You are a SQL assistant. Write a SQL query for the user's question.
}}
\caption{Stage 1 system prompt (complete).}
\label{fig:prompt_stage1}
\end{figure}

\begin{figure}[H]
\centering
\fbox{\parbox{0.94\linewidth}{\ttfamily\scriptsize\raggedright
    You are a PostgreSQL expert. Write a precise SELECT query for the given question.\\[0.9ex]
    === DATABASE RELATIONSHIPS ===\\
    The "crime\_incidents" table stores location as codes. Names live in lookup tables:\\
    ~~crime\_incidents."DISTRICT"~~~~~~~-> district\_names."district\_id"~~~~~~~~~~~~(name: "district\_name")\\
    ~~crime\_incidents."COMMUNITY\_AREA" -> community\_area\_names."community\_area\_id" (name: "community\_area\_name")\\
    ~~crime\_incidents."WARD"~~~~~~~~~~~-> ward\_names."ward\_id"~~~~~~~~~~~~~~~~~~~~(name: "ward\_name")\\[0.9ex]
    === CRITICAL RULES ===\\
    1. QUOTING IDENTIFIERS: Enclose column names that are upper-case in the schema in double\\
    ~~~quotes exactly as they appear, e.g. "YEAR", "DISTRICT", "CRIME\_TYPE", "INCIDENT\_DATE".\\[0.9ex]
    2. QUOTING STRING VALUES: ALWAYS use single quotes for string/text values:\\
    ~~~~~WHERE "CRIME\_TYPE" = 'THEFT'~~~~~~<- CORRECT\\
    ~~~~~WHERE "CRIME\_TYPE" = "THEFT"~~~~~~<- WRONG (double quotes = column name)\\
    ~~~Numeric values need no quotes: WHERE "YEAR" = 2020\\[0.9ex]
    3. BOOLEAN COLUMNS: "ARREST" and "DOMESTIC" are boolean columns. Filter them directly,\\
    ~~~e.g. WHERE "ARREST" = TRUE - never compare them to a string like = 'TRUE'.\\[0.9ex]
    4. DATES: "INCIDENT\_DATE" is a timestamp column.\\
    ~~~- For a month/quarter/year range, prefer a half-open range over BETWEEN:\\
    ~~~~~~~WHERE "INCIDENT\_DATE" >= '2024-07-01' AND "INCIDENT\_DATE" < '2024-10-01'\\
    ~~~- Use EXTRACT(YEAR | MONTH | HOUR FROM "INCIDENT\_DATE") to pull out a date part.\\[0.9ex]
    5. LOCATION NAMES: When the question refers to a district, community area or ward BY NAME,\\
    ~~~or asks for names in the output, JOIN the matching lookup table and use its name column.\\
    ~~~Never guess numeric codes for a named place.\\[0.9ex]
    6. MULTI-STEP LOGIC: Use CTEs (WITH clauses) for multi-step aggregations,\\
    ~~~year-over-year/period comparisons, or window-function pipelines. For window\\
    ~~~functions, always specify explicit PARTITION BY and ORDER BY clauses.\\[0.9ex]
    7. PERCENTAGES: Compute as ROUND(100.0 * numerator / NULLIF(denominator, 0), 2).\\[0.9ex]
    8. Only select columns relevant to the question. Do not add a LIMIT clause unless the\\
    ~~~question explicitly asks for a top-N or a specific number of rows.\\[0.9ex]
    9. NEVER use INSERT / UPDATE / DELETE / DROP / ALTER.\\[0.9ex]
    10. Return ONLY the SQL query, no markdown, no explanation.\\
    ~~~~If the question cannot be answered with a SELECT, respond: INVALID\_REQUEST
}}
\caption{Stage 2 system prompt (complete). Stages 3 and 4 start with exactly this text.}
\label{fig:prompt_stage2}
\end{figure}

\begin{figure}[H]
\centering
\fbox{\parbox{0.94\linewidth}{\ttfamily\scriptsize\raggedright
    === EXAMPLE ===\\
    Q: How many THEFT crimes were reported in 2024?\\
    SQL: SELECT COUNT(*) AS theft\_count FROM crime\_incidents WHERE "CRIME\_TYPE" = 'THEFT' AND "YEAR" = 2024;
}}
\caption{Text appended to the stage 2 prompt to form the stage 3 prompt.}
\label{fig:prompt_stage3}
\end{figure}

\begin{figure}[H]
\centering
\fbox{\parbox{0.94\linewidth}{\ttfamily\scriptsize\raggedright
    === EXAMPLES ===\\
    Q: How many THEFT crimes were reported in 2024?\\
    SQL: SELECT COUNT(*) AS theft\_count FROM crime\_incidents WHERE "CRIME\_TYPE" = 'THEFT' AND "YEAR" = 2024;\\[0.9ex]
    Q: What are the top 5 most common crime types?\\
    SQL: SELECT "CRIME\_TYPE", COUNT(*) AS crime\_count FROM crime\_incidents GROUP BY "CRIME\_TYPE" ORDER BY crime\_count DESC LIMIT 5;\\[0.9ex]
    Q: How many crimes were both domestic and resulted in an arrest?\\
    SQL: SELECT COUNT(*) AS domestic\_arrests FROM crime\_incidents WHERE "DOMESTIC" = TRUE AND "ARREST" = TRUE;\\[0.9ex]
    Q: In 2023, which months had more than 5000 total crimes?\\
    SQL: SELECT EXTRACT(MONTH FROM "INCIDENT\_DATE") AS month, COUNT(*) AS crime\_count FROM crime\_incidents WHERE "YEAR" = 2023 GROUP BY month HAVING COUNT(*) > 5000 ORDER BY month;
}}
\caption{Text appended to the stage 2 prompt to form the stage 4 prompt.}
\label{fig:prompt_stage4}
\end{figure}

Table~\ref{tab:hyperparameters} lists the settings that were held fixed in all experiments.

\begin{table}[H]
\centering
\small
\begin{tabularx}{\textwidth}{|l|X|X|}
\hline
\textbf{Parameter} & \textbf{Value} & \textbf{Rationale} \\
\hline
\textbf{Temperature} & 0 & Minimizes sampling variation \\
\hline
\textbf{Generation attempts} (\texttt{agent\_2}) & At most 3 in total & Bounds the number of retries and the added latency \\
\hline
\textbf{Relevance rule} (\texttt{agent\_2}) & At least one action verb and one data-context word & Decides relevance without a model call \\
\hline
\textbf{LLM Model} & openai/gpt-oss-120b & Open-weight model hosted on Groq \\
\hline
\end{tabularx}
\caption{Key settings held fixed in all experiments.}
\label{tab:hyperparameters}
\end{table}

\subsection{Evaluation Framework}
\label{sec:evaluation_framework}

\paragraph{Benchmark.} Performance was assessed on a hand-crafted benchmark of 100 natural language questions paired with ground-truth SQL, stratified as 30 \textbf{Easy} (single-table aggregations such as \texttt{COUNT}/\texttt{SUM}), 40 \textbf{Medium} (JOINs, \texttt{GROUP BY} with \texttt{HAVING}, temporal range analysis), and 30 \textbf{Hard} (window functions, CTEs, complex date arithmetic). Four questions are the worked examples embedded in the one-shot and few-shot prompts and are never scored, so every run scores \textbf{96 questions} (27 Easy, 39 Medium, 30 Hard). The reference SQL of every question was executed on the same database snapshot as the agents' queries; the largest reference result has 77 rows.

\paragraph{Experimental design.} No external baselines were run: a rule-based lexical matcher and a third-party zero-shot system were considered but not implemented, so no external comparison is reported. The comparison is internal: the two agents, \texttt{agent\_1} and \texttt{agent\_2}, built on the same model and schema, are each run with all four prompt stages. Every agent--stage pair is run \textbf{five times}, giving $2 \times 4 \times 5 = 40$ runs and 3,840 scored responses. The runs were collected sequentially, all \texttt{agent\_1} runs before the \texttt{agent\_2} runs (Section~\ref{sec:sq_artifact}), so the order in which runs were made is confounded with the agent. Although the temperature is 0, hosted inference is not bit-exact, so we report the mean and sample standard deviation (SD) over the five runs. This SD describes run-to-run variation on the same 96 questions; it is not an estimate of uncertainty about performance on other questions, databases or models.

\paragraph{Metrics.} Four metrics were used (Table~\ref{tab:metrics}).

\begin{table}[H]
\centering
\small
\begin{tabularx}{\textwidth}{lXl}
\toprule[1pt]
\textbf{Metric} & \textbf{Definition} & \textbf{Pre-set goal} \\
\midrule
\textbf{VSR} (Valid SQL Rate) & Percentage of questions for which the agent produced a query that executed without error; a question for which no query was produced counts as invalid & $\geq$ 95\% \\
\midrule
\textbf{EM} (normalized result match) & Percentage of questions whose result equals the reference result after the deterministic normalization described below; no model is involved & --- \\
\midrule
\textbf{EX} (judge-assisted execution accuracy) & Percentage of questions whose result matches the reference exactly (EM) or, for executed results that do not, is judged equivalent by an LLM judge & $\geq$ 85\% \\
\midrule
\textbf{SQ} (Synthesis Quality) & Mean Likert score (1--5) assigned by the LLM judge to the agent's plain-English answer, judged against the rows its own query returned & $\geq$ 4.0 \\
\bottomrule[1pt]
\end{tabularx}
\caption{Evaluation metrics and pre-set goals. EM is deterministic; EX and SQ depend in part on an LLM judge. A question for which no query was produced counts as invalid for VSR and incorrect for EM and EX. Goals were set before development.}
\label{tab:metrics}
\end{table}

\begin{table}[H]
\centering
\small
\begin{tabularx}{\textwidth}{cX}
\toprule[1pt]
\textbf{Score} & \textbf{Criterion} \\
\midrule
5 & Fully faithful to the result rows and directly answers the question; numbers and names are correct, nothing important is omitted, no facts are invented. \\
\midrule
4 & Faithful and answers the question, with a minor flaw (slight omission, awkward framing, unnecessary hedging, small rounding looseness). \\
\midrule
3 & Partly useful: answers only part of the question, ignores part of the result, is vague, or contains a noticeable misreading that does not flip the answer. \\
\midrule
2 & Substantially misleading: misstates the results, states a conclusion the rows do not support, or answers a different question. \\
\midrule
1 & Unusable or wrong: fabricated numbers, contradicts the rows, an empty or error response, or reports ``no results'' or ``out of scope'' when the question is in scope and data exist. \\
\bottomrule[1pt]
\end{tabularx}
\caption{Rubric for Synthesis Quality (SQ). The judge scores the answer against the query result it is shown, not against the reference. An empty result with an answer saying nothing was found is faithful; if the query failed or no query ran, an answer saying ``no results found'' or ``out of scope'' scores 1.}
\label{tab:sq_rubric}
\end{table}

\paragraph{Scoring procedure.} Evaluation proceeded through four stages.
\begin{enumerate}
    \item \textbf{Query execution.} The agent's SQL and the reference SQL are re-executed against the PostgreSQL database in a read-only session with a 120-second per-query timeout. Each distinct SQL string is executed once and reused. Queries that raise an error, and questions for which no query was produced, are invalid and give VSR\,=\,0.
    \item \textbf{Result comparison (EM).} A result equal to the reference after normalization counts as an exact match. Normalization ignores row order, column names, letter case and surrounding whitespace, and rounds numbers to two decimals; column order and extra or missing columns are not ignored and numeric strings are read as numbers (for example, "25.00" equals 25, duplicate rows are compared as a multiset, not a set. Results are compared on at most their first 1,000 rows, which no reference result exceeds.
    \item \textbf{Judge grading (EX and SQ).} Every result from a query that ran but did not match exactly is graded for equivalence by an LLM judge, which decides EX for those responses. The judge also scores every answer text for SQ. Responses with no executable SQL get EX\,=\,0, and answers to questions where no query ran, or the query failed and the answer reports ``no results'', get SQ\,=\,1 by rule, both without a judge call.
    \item \textbf{Error categorization.} Every incorrect answer is assigned to one of three failure types (no query produced, query failed to execute, query executed with a wrong result) and to a difficulty tier.
\end{enumerate}

\paragraph{What the comparison does and does not establish.} EM and EX compare query \emph{results} on one database snapshot; neither establishes that the generated SQL is semantically equivalent to the reference query, and a query can match on this snapshot for reasons that would not hold on other data. The normalization is deliberately permissive. Because row order is ignored, a ranking returned in the wrong order can still count as an exact match, which matters for questions whose answer is an ordered list. Rounding to two decimals can hide small numerical differences, for example in rates and averages. Because column order and extra or missing columns are not ignored, a result with its columns in another order, or with an additional supporting column, fails EM and is passed to the judge. A generated result longer than 1,000 rows is compared only on its first 1,000 rows; since no reference result exceeds 77 rows, such a result cannot match exactly. The judge never sees a full result: it is shown at most the first 20 rows and 700 characters of both results and is not told the total number of rows, so differences beyond that portion cannot be detected.No agent result in our runs exceeded 77 rows. The judge's decisions have not been audited manually, and no human agreement statistics are available.

\subsection{Implementation Details}

The system is written in Python (the code requires version 3.10 or newer; the interpreter version used for the evaluation runs is not recorded here) using LangGraph for the agent graphs, LangChain for the model and database interfaces, the Groq SDK for inference, SQLAlchemy and psycopg2 for database access, and the Anthropic SDK for the judge (Table~\ref{tab:implementation}). Weekly data ingestion is orchestrated by Apache Airflow. For the benchmark, the database runs as a Docker container built from the published image: PostgreSQL~15 with the 8,643,493-row snapshot loaded at build time, which needs about 6.5~GB of disk and at least 4~GB of memory to run locally.

\begin{table}[H]
\centering
\small
\begin{tabularx}{\textwidth}{lX}
\toprule[1pt]
\textbf{Component} & \textbf{Version / setting} \\
\midrule
Agent framework & LangGraph 1.0.10, LangChain Core 1.2.17, LangChain Community 0.4.1, LangChain Groq 1.1.2 \\
\midrule
LLM inference & \texttt{openai/gpt-oss-120b} on Groq (Groq SDK 0.37.1), temperature 0, developer tier \\
\midrule
Database access & SQLAlchemy 2.0.48, psycopg2-binary 2.9.11 \\
\midrule
Database & PostgreSQL 15 in Docker (\texttt{vigneshwarr/chicago-crime-db}), read-only role for the agents \\
\midrule
Judge & \texttt{anthropic/claude-haiku-5-5} via the Anthropic Python SDK \\
\midrule
Parallelism & 8 questions in parallel within a run \\
\midrule
Schema shown to the model & Four tables: \texttt{crime\_incidents}, \texttt{district\_names}, \texttt{community\_area\_names}, \texttt{ward\_names} \\
\bottomrule[1pt]
\end{tabularx}
\caption{Software and configuration used in the experiments.}
\label{tab:implementation}
\end{table}

The agents connect through the database user \texttt{ai\_agent}, which has read-only privileges, so that neither agent can run data-modifying queries.

\section{Results}
\label{sec:results}

All figures are the mean and sample standard deviation over five runs of 96 questions each (Section~\ref{sec:evaluation_framework}). A question for which an agent produced no query counts as invalid for VSR and as incorrect for EM and EX.

\subsection{Overall Performance}

Table~\ref{tab:main_results} reports VSR, EX and SQ for each agent and prompt stage; EM is reported in Table~\ref{tab:judge_share}.

\begin{table}[H]
\centering
\small
\begin{tabular}{llrrr}
\toprule
\textbf{Agent} & \textbf{Prompt stage} & \textbf{VSR (\%)} & \textbf{EX (\%)} & \textbf{SQ (1--5)} \\
\midrule
\texttt{agent\_1} & 1 simple    & $67.1 \pm 2.2$ & $62.1 \pm 2.5$ & $2.33 \pm 0.04$ \\
\texttt{agent\_1} & 2 rules     & $99.0 \pm 1.0$ & $93.1 \pm 1.2$ & $2.90 \pm 0.08$ \\
\texttt{agent\_1} & 3 one-shot  & $98.5 \pm 0.6$ & $91.9 \pm 0.5$ & $2.91 \pm 0.04$ \\
\texttt{agent\_1} & 4 few-shot  & $97.9 \pm 0.7$ & $91.5 \pm 1.4$ & $2.88 \pm 0.04$ \\
\midrule
\texttt{agent\_2} & 1 simple    & $91.7 \pm 0.0$ & $85.0 \pm 0.6$ & $2.69 \pm 0.04$ \\
\texttt{agent\_2} & 2 rules     & $91.7 \pm 0.0$ & $86.2 \pm 0.5$ & $3.52 \pm 0.70$ \\
\texttt{agent\_2} & 3 one-shot  & $91.7 \pm 0.0$ & $86.7 \pm 0.9$ & $4.02 \pm 0.02$ \\
\texttt{agent\_2} & 4 few-shot  & $91.7 \pm 0.0$ & $87.5 \pm 0.7$ & $4.05 \pm 0.04$ \\
\bottomrule
\end{tabular}
\caption{Performance of the two agents at each prompt stage (mean $\pm$ SD over five runs, 96 questions per run). EX is judge-assisted. Pre-set goals: VSR $\geq$ 95\%, EX $\geq$ 85\%, SQ $\geq$ 4.0. The SQ column is confounded with run order and should not be used to compare agents or stages (Section~\ref{sec:sq_artifact}).}
\label{tab:main_results}
\end{table}

\texttt{agent\_1} with any rule-based prompt (stages 2--4) meets the VSR goal (97.9--99.0\%) and the EX goal (91.5--93.1\%). \texttt{agent\_2} meets the EX goal at every stage (stage 1 only just: $85.0 \pm 0.6$\%) but misses the VSR goal at 91.7\%, because its relevance checker rejects 8 of the 96 questions (Section~\ref{sec:retry}). Only \texttt{agent\_2} at stages 3 and 4 reaches the SQ goal, but those SQ values are confounded with run order (Section~\ref{sec:sq_artifact}), so no configuration can be said to meet all three goals. With the bare stage-1 prompt, neither agent reaches the VSR goal.

\subsection{Effect of the Prompt Stage}

Explicit rules are the large step, and examples add little. For \texttt{agent\_1}, moving from the bare prompt to explicit rules raises VSR from 67.1\% to 99.0\% and EX from 62.1\% to 93.1\%. One-shot and few-shot examples do not raise EX (91.9\% and 91.5\%, against 93.1\% with rules only); these differences of 1.2 and 1.6 points are of the same order as the run-to-run standard deviations (0.5--1.4 points). No significance tests were performed, and none of the differences reported in this section should be read as statistically established.

The examples change the form of the result more than its correctness. The share of results that match the reference exactly (EM, without the judge) rises for \texttt{agent\_1} from 77.3\% (rules) to 79.0\% (one-shot) to 83.5\% (few-shot), while EX stays flat (Table~\ref{tab:judge_share}). Because the examples are themselves reference queries from the benchmark, this is consistent with the examples steering the output toward the reference style (column choice, ordering, scale) rather than making the agent more often correct.

The bare stage-1 prompt fails mainly through identifier quoting. 158 of \texttt{agent\_1}'s 480 stage-1 queries (33\%) did not execute; the model writes upper-case identifiers unquoted and PostgreSQL folds them to lower case, giving errors such as \texttt{column "incident\_date" does not exist} (38 times) and \texttt{ci.community\_area does not exist} (25). Without a retry, \texttt{agent\_1} cannot recover. At this stage \texttt{agent\_2}'s VSR and EX are 24.6 and 22.9 points higher than \texttt{agent\_1}'s (91.7\% against 67.1\%, and 85.0\% against 62.1\%), and 146 of its 440 generated queries failed the first validation and were repaired (Section~\ref{sec:retry}). The difference is consistent with the retry loop compensating for the quoting failures, although the two agents also differ in relevance filtering. Stage 1 is therefore not a ``no prompt engineering'' floor that is comparable between the two agents.

\subsection{Per-Tier Trends}

Table~\ref{tab:tier_results} breaks VSR and EX down by difficulty tier. The tiers are small (27--39 questions), so differences of a few points correspond to one or two questions.

\begin{table}[H]
\centering
\small
\setlength{\tabcolsep}{4pt}
\begin{tabular}{llrrrrrr}
\toprule
 & & \multicolumn{2}{c}{\textbf{Easy (27)}} & \multicolumn{2}{c}{\textbf{Medium (39)}} & \multicolumn{2}{c}{\textbf{Hard (30)}} \\
\midrule
\textbf{Agent} & \textbf{Stage} & VSR & EX & VSR & EX & VSR & EX \\
\midrule
\texttt{agent\_1} & 1 simple   &  63 &  63 &  68 &  63 &  69 &  61 \\
\texttt{agent\_1} & 2 rules    & 100 & 100 &  99 &  90 &  98 &  91 \\
\texttt{agent\_1} & 3 one-shot & 100 & 100 &  98 &  89 &  98 &  88 \\
\texttt{agent\_1} & 4 few-shot & 100 & 100 &  99 &  92 &  95 &  83 \\
\midrule
\texttt{agent\_2} & 1 simple   &  96 &  96 &  87 &  82 &  93 &  79 \\
\texttt{agent\_2} & 2 rules    &  96 &  96 &  87 &  82 &  93 &  83 \\
\texttt{agent\_2} & 3 one-shot &  96 &  96 &  87 &  82 &  93 &  85 \\
\texttt{agent\_2} & 4 few-shot &  96 &  96 &  87 &  82 &  93 &  87 \\
\bottomrule
\end{tabular}
\caption{VSR and EX (\%, mean over five runs) by difficulty tier. Tier sizes are given after removing the four reserved questions. The \texttt{agent\_2} ceilings (96, 87 and 93\% VSR) reflect the questions rejected by its relevance checker (1 Easy, 5 Medium, 2 Hard). In these tiers one question corresponds to 2.6--3.7 points. SQ is not broken down by tier because of the confound described in Section~\ref{sec:sq_artifact}.}
\label{tab:tier_results}
\end{table}

\paragraph{Easy tier.} With rule-based prompts, \texttt{agent\_1} answers every Easy question correctly (VSR and EX 100\%). \texttt{agent\_2} is capped at 96\% because one Easy question (``List all distinct location risk categories'') is rejected by its relevance checker in every run.

\paragraph{Medium tier.} \texttt{agent\_1} reaches 89--92\% EX. \texttt{agent\_2} is the weakest here (VSR 87\%, EX 82\%), because 5 of the 8 questions its relevance checker rejects are Medium questions.

\paragraph{Hard tier.} Window functions, CTEs and multi-step logic are the hardest queries. With rule-based prompts EX is 83--91\% for \texttt{agent\_1} and 83--87\% for \texttt{agent\_2}. \texttt{agent\_2}'s Hard-tier EX rises across the four stages from 79\% to 87\%, while \texttt{agent\_1}'s declines across stages 2--4 from 91\% to 83\% and its VSR falls from 98\% to 95\%. These are differences of two to three questions out of 30.

\subsection{What the Validation-and-Retry Loop Does}
\label{sec:retry}

The two agents differ in two components at once: the relevance gate and the validate-and-retry loop. Table~\ref{tab:agent_comparison} summarizes stages 2--4, the realistic configurations, with explicit denominators.

\begin{table}[H]
\centering
\small
\begin{tabular}{lrr}
\toprule
 & \texttt{agent\_1} & \texttt{agent\_2} \\
\midrule
Questions that received a query & 96 / 96 & 88 / 96 \\
Queries produced that executed without error & 98.5\% (of 96) & 100.0\% (of 88) \\
EX among questions that received a query & 92.2\% (of 96) & 94.7\% (of 88) \\
EX over all 96 questions & 92.2\% & 86.8\% \\
\bottomrule
\end{tabular}
\caption{Comparison of the two agents, stages 2--4 (mean of the three stages). The second and third rows use different question sets for the two agents and are not directly comparable; only the last row uses the same 96 questions for both.}
\label{tab:agent_comparison}
\end{table}

\paragraph{The retry loop repaired every observed first-pass failure.} In stages 2--4, 14 of \texttt{agent\_2}'s 1,320 generated queries (1.1\%) failed \texttt{EXPLAIN} on the first attempt; at stage 1, 146 of its 440 queries (33.2\%) did. In all 160 cases a later attempt passed validation within the attempt limit, and none used up all three attempts; every query \texttt{agent\_2} produced executed without error. Here ``repaired'' means that the query passed validation and executed, not that it returned the correct result: the manuscript's records do not report how many of the 160 repaired queries were also correct. For comparison, 22 of \texttt{agent\_1}'s 1,440 queries at stages 2--4 (1.5\% on average; 1.0\%, 1.5\% and 2.1\% at stages 2, 3 and 4) failed to execute; 13 of these are a single error (\texttt{round(double precision, integer) does not exist}) that PostgreSQL raises before execution and that \texttt{EXPLAIN}-based validation would therefore detect. Whether a retry would then have corrected these queries was not tested for \texttt{agent\_1}.

\paragraph{Accuracy on the accepted questions cannot be compared directly.} \texttt{agent\_2}'s 94.7\% is computed on the 88 questions its relevance checker accepted, whereas \texttt{agent\_1}'s 92.2\% is computed on all 96, so the 2.5-point difference is not evidence of a benefit from retry. The aggregate results bound \texttt{agent\_1}'s accuracy on the same 88 questions: it answered on average 88.5 of 96 questions correctly, so its EX on the 88 accepted questions lies between 91.5\% (if it answered all eight rejected questions correctly) and 100\%. \texttt{agent\_2}'s 94.7\% falls inside this interval; resolving the comparison requires per-question results restricted to the 88 accepted questions.

\paragraph{How the relevance checker works.} The relevance checker of \texttt{agent\_2} is a rule-based filter that makes no model call to decide. It lower-cases the question and applies three tests in order. (i)~A question that asks about the database itself in general terms (for example ``what is the database'') is rejected. (ii)~A question that matches an out-of-scope pattern (arithmetic, ``what is the meaning'', ``define'', ``explain'', ``tell me about'', and questions starting ``who'', ``when'' or ``why'') is rejected. (iii)~Any other question is accepted only if it contains \emph{both} an action phrase and a data-context word. The action phrases are ``how many'', ``how often'', ``what are the'', ``list'', ``show'', ``find'', ``get'', ``count'', ``total'', ``sum'', ``average'', ``top'', ``most'', ``least'', ``rank'', ``filter'', ``between'', ``after'', ``before'', ``during'', ``in the year'' and ``percentage''. The data-context words are \emph{crime}, \emph{crimes}, \emph{chicago}, \emph{report}, \emph{reported}, \emph{database}, \emph{table}, \emph{records}, \emph{data}, \emph{district}, \emph{year}, \emph{date}, \emph{type}, \emph{arrest}, \emph{victim}, \emph{incidents}, \emph{cases}, \emph{statistics}, \emph{breakdown} and \emph{distribution}. Both lists are matched as substrings, and neither includes comparatives such as ``highest'' or ``more than''. For a rejected question no query is generated, and the run ends with a short out-of-scope reply that the language model writes from a fixed instruction. The rejection is decided by the rules alone.

\paragraph{Why the eight questions are rejected.} None of the eight questions matches the first two tests. All are rejected by the third, because each lacks either an action phrase or a data-context word (Table~\ref{tab:rejected}). Seven contain data words but no action phrase: they ask ``Which \ldots'' or ``What is the \ldots'', and the list contains ``what are the'' but not ``what is the'', ``which'', ``highest'' or ``more than''. Question~28 is the reverse: it begins with ``List'' and so has an action phrase, but ``location risk categories'' contains none of the data-context words.

\begin{table}[H]
\centering
\small
\begin{tabularx}{\textwidth}{rXllX}
\toprule[1pt]
\textbf{ID} & \textbf{Question} & \textbf{Action phrase} & \textbf{Data word} & \textbf{Why it is rejected} \\
\midrule
28 & List all distinct location risk categories. & list & none & No data-context word. \\
\midrule
35 & What is the arrest rate for each crime category? & none & arrest, crime & ``What is the'' is not an action phrase. \\
\midrule
37 & Which 5 districts (by name) have the highest number of arrests? & none & arrest, district & ``Which'' and ``highest number'' are not action phrases. \\
\midrule
42 & Which districts (by name) had more than 20000 crimes in 2024? & none & crime, district & ``Which'' and ``more than'' are not action phrases. \\
\midrule
44 & Which year had the highest number of homicide offenses? & none & year & ``Which'' and ``highest number'' are not action phrases. \\
\midrule
62 & Which community areas (by name) had more than 5000 property crimes? & none & crime & ``Which'' and ``more than'' are not action phrases. \\
\midrule
95 & Which crime clearance bucket has the highest proportion of crimes resulting in an arrest? & none & crime, arrest & ``Which'' and ``highest proportion'' are not action phrases. \\
\midrule
96 & Which ward (by name) had the single highest number of weapon crimes in any one year? & none & crime, year & ``Which'' and ``highest number'' are not action phrases. \\
\bottomrule[1pt]
\end{tabularx}
\caption{The eight questions rejected by the relevance checker in all 20 runs of \texttt{agent\_2}, with the action phrase and data-context words each contains. Seven fail the action-phrase test and one (ID~28) fails the data-context test.}
\label{tab:rejected}
\end{table}

With explicit prompt rules and their one-shot and few-shot extensions (stages 2--4), the simple agent achieved higher judge-assisted query-result accuracy and VSR than the full agent (EX 92.2\% against 86.8\%, VSR 98.5\% against 91.7\%, both over all 96 questions); with the bare stage-1 prompt, the full agent performed better (EX 85.0\% against 62.1\%, VSR 91.7\% against 67.1\%). These figures concern query results; they do not measure the correctness of the final natural-language answers, whose SQ scores are confounded (Section~\ref{sec:sq_artifact}). Because the agents differ in two components, this difference is a joint effect and does not isolate either one. A configuration with validation and retry but without the relevance checker was not tested.

\subsection{Failure Analysis}

Table~\ref{tab:failures} counts, per run, the questions with EX\,=\,0 by failure type.

\begin{table}[H]
\centering
\small
\begin{tabular}{llrrr}
\toprule
\textbf{Agent} & \textbf{Stage} & \textbf{No query} & \textbf{Failed to execute} & \textbf{Wrong result} \\
\midrule
\texttt{agent\_1} & 1 simple   & 0   & 31.6 & 4.8 \\
\texttt{agent\_1} & 2 rules    & 0   & 1.0  & 5.6 \\
\texttt{agent\_1} & 3 one-shot & 0   & 1.4  & 6.4 \\
\texttt{agent\_1} & 4 few-shot & 0   & 2.0  & 6.2 \\
\midrule
\texttt{agent\_2} & 1 simple   & 8.0 & 0    & 6.4 \\
\texttt{agent\_2} & 2 rules    & 8.0 & 0    & 5.2 \\
\texttt{agent\_2} & 3 one-shot & 8.0 & 0    & 4.8 \\
\texttt{agent\_2} & 4 few-shot & 8.0 & 0    & 4.0 \\
\bottomrule
\end{tabular}
\caption{Mean number of questions per run (of 96) with EX\,=\,0, by failure type. For \texttt{agent\_2}, the 8.0 questions without a query are those rejected by the relevance checker. ``Wrong result'' counts queries that executed but were neither an exact match nor judged equivalent.}
\label{tab:failures}
\end{table}

Queries that execute but return a wrong result remain at 4--6 per run (about 5\% of questions) in every configuration of both agents; neither the prompt stage nor the retry loop removes them, because the retry loop only reacts to queries that fail validation. The judge's rejection reasons illustrate the causes: a different interpretation of an ambiguous question (for example, ``how many distinct districts reported at least one crime in each year'' answered with one overall figure instead of a per-year count), different logic for ``districts with zero homicide offenses'' (23 reported against 4 in the reference), and result sets that include or omit a qualifying community area or ward.

\subsection{Contribution of the Judge to EX}

Table~\ref{tab:judge_share} shows how much of EX comes from the judge. Overall, 2,824 of the 3,840 responses (73.5\%) match the reference exactly (EM), and the judge classified a further 459 as equivalent (12.0 points), giving 85.5\% EX. Of the 676 executed responses that did not match exactly, the judge accepted 459 and rejected 217 (106 and 42 of 148 distinct query--reference pairs). The judge's stated reasons for these classifications describe differences such as an extra or missing supporting column, another row order, or the same quantity in another form (weekday names instead of 0--6, a full date instead of a month number). Whether every accepted result is in fact equivalent has not been checked. These acceptances have not been audited manually, so the 12.0-point difference between EM and EX rests on unvalidated LLM decisions; EM is the deterministic, lower figure.

\begin{table}[H]
\centering
\small
\begin{tabular}{llrr}
\toprule
\textbf{Agent} & \textbf{Stage} & \textbf{EM (\%)} & \textbf{EX (\%)} \\
\midrule
\texttt{agent\_1} & 1 simple   & 52.9 & 62.1 \\
\texttt{agent\_1} & 2 rules    & 77.3 & 93.1 \\
\texttt{agent\_1} & 3 one-shot & 79.0 & 91.9 \\
\texttt{agent\_1} & 4 few-shot & 83.5 & 91.5 \\
\midrule
\texttt{agent\_2} & 1 simple   & 72.1 & 85.0 \\
\texttt{agent\_2} & 2 rules    & 71.9 & 86.2 \\
\texttt{agent\_2} & 3 one-shot & 73.3 & 86.7 \\
\texttt{agent\_2} & 4 few-shot & 78.3 & 87.5 \\
\bottomrule
\end{tabular}
\caption{Normalized result match (EM) and judge-assisted execution accuracy (EX), mean over five runs. EM ignores row order, column names, letter case and whitespace and rounds numbers to two decimals; EX also counts executed results that the LLM judge classified as equivalent. The judge's decisions were not audited manually, so the EX--EM difference rests on unvalidated judgments.}
\label{tab:judge_share}
\end{table}

\subsection{Synthesis Quality}
\label{sec:sq_artifact}

SQ scores the plain-English answer against the rows the agent's own query returned. In Table~\ref{tab:main_results} it is low for \texttt{agent\_1} (2.3--2.9) and rises from 2.7 to 4.1 across the stages of \texttt{agent\_2}. These differences cannot be attributed to the agents' design: they coincide with a number-copying error in the answer synthesizer whose frequency changed with the order in which runs were made.

\paragraph{The error.} The synthesizer sometimes changes digits when it restates a number from the result. For example, a query returned 2,158,897 arrests and the answer said 2,158,806. The judge scored such answers 1 or 2. We counted the questions whose query returns a single whole number of at least 1,000. Across the 849 answers to these questions, only 399 (47\%) repeat the number exactly.

\paragraph{When it happened.} Table~\ref{tab:digit_copy} shows that the frequency of copying errors changed across sequential runs, but run order, agent configuration and prompt stage were not independently varied, so their contributions cannot be separated. In all of \texttt{agent\_1}'s runs and in the first part of \texttt{agent\_2}'s runs, only 17--21\% of answers repeat the number exactly. From the third run of \texttt{agent\_2} at stage 2 onward, every answer does. The pipeline code was not changed during the runs. One possible explanation is a provider-side change in how the model was served. This is unverified, not an established cause, and the possible effects of any such change cannot be assumed to have been restricted to answer synthesis; the available records do not establish whether SQL generation or the other reported outcomes were also affected.

\begin{table}[H]
\centering
\small
\begin{tabular}{lrr}
\toprule
\textbf{Runs} & \textbf{Exact copies} & \textbf{Share} \\
\midrule
\texttt{agent\_1}, stage 1                                      & 11 of 65   & 17\% \\
\texttt{agent\_1}, stages 2--4                                  & 71 of 341  & 21\% \\
\texttt{agent\_2}, stage 1                                      & 20 of 110  & 18\% \\
\texttt{agent\_2}, stage 2, runs 1--2                           & 8 of 44    & 18\% \\
\texttt{agent\_2}, stage 2 runs 3--5, and stages 3--4           & 289 of 289 & 100\% \\
\bottomrule
\end{tabular}
\caption{Answers to single-number questions (results of at least 1,000) that repeat the query's number exactly, pooled over runs and grouped by when the runs were made (\texttt{agent\_1} first, then \texttt{agent\_2}). Run order, agent and prompt stage were not independently varied, so the cause of the change cannot be identified.}
\label{tab:digit_copy}
\end{table}

\paragraph{Consequence.} The pattern matches the SQ column. \texttt{agent\_2} at stage 2 has a standard deviation of 0.70 because its first two runs average 2.75 and its last three 4.03. The higher SQ of \texttt{agent\_2} at stages 3 and 4 coincides with when those runs were made and cannot be attributed to the prompt or the validation loop. SQ therefore cannot be used to compare the agents or the prompt stages here. The copying errors themselves remain descriptive evidence that answer synthesis is a separate source of error from SQL generation.

\section{Discussion}
\label{sec:interpretation}

\subsection{Four Distinct Outcomes}

The results separate four outcomes that a single accuracy figure would merge.

\paragraph{Executable SQL.} For \texttt{agent\_1}, the explicit rules of the stage-2 prompt were the main driver of validity, raising VSR from 67.1\% to 99.0\%. For \texttt{agent\_2}, the validation-and-retry loop ensured that every query it produced executed, at every prompt stage, and it repaired all 160 observed first-pass failures. When first-pass failures are frequent, as with the bare stage-1 prompt, repair matters most.

\paragraph{The correct result.} Executable SQL is not the same as a correct result. Four to six questions per run returned a wrong result in every configuration, mostly through a different reading of an ambiguous question or different query logic, and the retry loop cannot address these because they pass validation. The 12-point gap between EM and EX further shows that whether a result counts as correct depends on the comparison procedure.

\paragraph{An accurate summary.} A correct result can still be summarized incorrectly. Only 47\% of answers to single-number questions repeated the number exactly, and the frequency of this error changed over the course of the experiment. Answer synthesis is therefore a distinct source of error, not a step that can be assumed correct once the SQL is. The synthesizer also sees only the first ten result rows, so answers to questions with longer results are written from partial results.

\paragraph{Performance after filtering and retry.} Added components can introduce failures of their own. At stages 2--4, first-pass validation failures affected 1.1\% of \texttt{agent\_2}'s generated queries, and execution failures affected 1.5\% of \texttt{agent\_1}'s queries on average, so the room for repair to improve query-result accuracy on this benchmark was small. The relevance gate rejected eight of 96 questions in every run, imposing a 91.7\% ceiling on overall VSR and accuracy; how much accuracy removing the gate would recover was not measured. With the bare stage-1 prompt, where first-pass failures were frequent, the full agent performed better. In practice this argues for measuring the false-rejection rate of an input filter as carefully as the accuracy of the query generator, and for evaluating each component by its effect on end-to-end answers rather than on its own success rate.

\subsection{Prompting}

Explicit, schema-specific rules produced the large improvement; one-shot and few-shot examples raised exact matches but not judge-assisted accuracy. On this benchmark, examples mainly made results look more like the reference results. Because the stage-1 prompt fails mostly on identifier quoting, the size of the stage-1 to stage-2 improvement is specific to a schema with upper-case identifiers.

\subsection{Limitations}

\begin{itemize}
  \item \textbf{Scope:} All results come from one model (\texttt{openai/gpt-oss-120b}), one database and a benchmark of 96 scored questions written by the authors. Spider \cite{yu2018spider} and BIRD \cite{li2023bird} were not run and no external baseline was executed, so the findings describe this agent on this dataset, not its standing against prior systems.
  \item \textbf{Confounded comparison:} The agents differ in both relevance filtering and validation with retry, and no configuration with only one of these components was run, so neither component's effect is isolated.
  \item \textbf{Run order:} Runs were collected sequentially, \texttt{agent\_1} before \texttt{agent\_2}. Run order is therefore confounded with the agent and, within \texttt{agent\_2}, with the prompt stage. The change in number copying across sequential runs suggests that service behavior may have varied, but because run order, agent and prompt stage were not independently varied, its cause cannot be identified, and whether any such variation also affected SQL generation is unknown.
  \item \textbf{Variation and uncertainty:} The reported standard deviations describe variation across five runs on the same questions. They do not quantify uncertainty about performance on other questions, and the difficulty tiers contain only 27--39 questions each.
  \item \textbf{Judge-dependent metrics:} EX relies on an LLM judge for 676 non-matching responses and SQ entirely on the judge. The judge's decisions were not audited manually and no agreement statistics are available. SQ is additionally confounded with run order.
  \item \textbf{Permissive comparison:} EM ignores row order and rounds to two decimals, so it can accept results that differ in ordering or precision; it compares results on one snapshot and does not establish SQL equivalence.
  \item \textbf{Partial results for synthesis:} The answer synthesizer receives only the first ten rows, while reference results contain up to 77 rows.
  \item \textbf{Imputed data:} Missing crime categories were replaced with OTHER OFFENSE, and missing coordinates and dates were imputed (Table~\ref{tab:null_handling}). Counts by category, location or date therefore include imputed values; for example, a count of OTHER OFFENSE incidents includes records whose category was missing, and imputed coordinates place records at block or beat averages rather than recorded locations. Both reference and generated queries run on the same imputed data, so this affects the interpretation of the answers rather than the comparison between them.
  \item \textbf{Relevance rule:} The relevance checker is a fixed keyword rule. Its false-rejection rate on other phrasings, and its effect on genuinely out-of-scope questions, were not measured.
  \item \textbf{No user study:} Accessibility and usability for non-technical users were not evaluated; the study measures query and answer correctness only.
\end{itemize}

\section{Future Work}

The following are proposals, not demonstrated outcomes.

\paragraph{Component ablations and controlled repetition.} The most direct next step is to run \texttt{agent\_2} without the relevance checker and \texttt{agent\_1} with validation and retry only, with runs interleaved or randomized across agents and stages, and to repeat the 27 runs affected by the number-copying error (all of \texttt{agent\_1}, \texttt{agent\_2} stage 1, and \texttt{agent\_2} stage 2 runs 1--2) so that SQ can be compared.

\paragraph{Relevance checking.} The keyword rule could be replaced by a classifier or by embedding similarity to a set of annotated in-domain questions, using a sentence encoder such as SBERT \cite{reimers2019sbert}. Any replacement should be evaluated by its false-rejection and false-acceptance rates on held-out phrasings.

\paragraph{Retrieval for schema and examples.} Retrieval-augmented generation \cite{lewis2020rag} offers one route to selecting the relevant schema columns and in-context examples for each question instead of supplying the full schema and fixed examples. Prior work has explored retrieving structurally similar examples for Text-to-SQL prompting \cite{guo2023prompting}, and a vector index such as FAISS \cite{johnson2017faiss} could support this. Whether retrieval improves accuracy on this benchmark would need to be measured.

\paragraph{Answer verification and evaluation.} Because answer synthesis introduced numerical errors, a deterministic check that every number in an answer appears in the result rows would directly target the observed failure. Embedding-based measures such as BERTScore \cite{zhang2020bertscore} capture similarity of wording and are not sufficient to verify numerical or factual correctness. For EX, a manual audit of a sample of judge decisions, and stricter comparison options for questions where order or precision matters, would make the reported accuracy more reliable. Evaluating additional models and a subset of a public benchmark would test how far the findings generalize.

\section{Conclusion}

We evaluated a LangGraph Text-to-SQL agent on the Chicago Crime database by comparing a simple pipeline with a full pipeline that adds a relevance checker and validation with retry, across four prompt configurations and five repeated runs. Explicit prompt rules made nearly all generated SQL executable, and the retry loop repaired every observed validation failure. With explicit prompt rules and their one-shot and few-shot extensions, the simpler pipeline achieved higher judge-assisted query-result accuracy; with the bare prompt, the full pipeline performed better. The relevance gate rejected eight of 96 questions in every run, imposing a 91.7\% ceiling on the full agent's overall VSR and accuracy; the gain from removing it was not measured. Query-result accuracy is not the same as the correctness of the final answer: correct results were not always summarized correctly, and the confounded SQ scores do not show which agent gave users better answers. Within the limits of a single model, a single database and an author-built benchmark, these results show that executable SQL, correct results, accurate summaries and end-to-end performance should be measured separately, and that each added component should be judged by its effect on the answers users receive.

\section*{Statements and Declarations}

\noindent\textbf{Funding.} No funding was received for this study.
\par\vspace{8pt}

\noindent\textbf{Competing interests.} The authors have no conflict of interest to declare.
\par\vspace{8pt}

\noindent\textbf{Ethics approval and consent.} Not applicable: the study uses publicly released administrative data and involves no human participants.
\par\vspace{8pt}

\noindent\textbf{Data and code availability.} The database snapshot used in all experiments is available as the Docker image \texttt{vigneshwarr/chicago-crime-db}. The benchmark questions with reference SQL, the evaluation code, the image build files, the per-run outputs and the judge decisions are available at \url{https://github.com/Vigneshwarr3/chicago-text2sql}.
\par\vspace{8pt}

\noindent\textbf{Author contributions.}
\textbf{Vigneshwar Ravi Rao and Rupesh Swarnakar:}
Writing -- review \& editing, Writing -- original draft, Methodology, Investigation, Data curation, and Formal analysis.
\textbf{Fayeq Jeelani Syed:}
Conceptualization, Supervision, Investigation, Data curation, and Writing -- review \& editing.
\par\vspace{8pt}

\noindent\textbf{Use of generative AI.} ChatGPT (OpenAI) was used to improve the manuscript's language, organization and flow, adjust section lengths, and assist with creating and formatting tables. Claude (Anthropic) was used to check the internal consistency of the reported figures and to revise the text accordingly. The authors reviewed and edited the AI-assisted content and take full responsibility for the final manuscript. This declaration concerns manuscript preparation; the use of language models within the Text-to-SQL system and as an evaluation judge is described in the Methods section.

\bibliographystyle{IEEEtran}
\bibliography{sample}

\end{document}